\documentclass{article}

\usepackage[nonatbib,final]{neurips_2022}

\usepackage[utf8]{inputenc}   
\usepackage[T1]{fontenc}      
\usepackage[backref=page]{hyperref}         
\usepackage{url}              
\usepackage{booktabs}         
\usepackage{amsmath}          
\usepackage{amssymb}          
\usepackage{nicefrac}         
\usepackage{microtype}        
\usepackage{graphicx}         
\usepackage[numbers]{natbib}  
\usepackage{xcolor}           
\usepackage{xspace}           
\usepackage{wrapfig}          
\usepackage{subcaption}       
\usepackage{tikz}             
\usetikzlibrary{arrows.meta,positioning,calc}
\usepackage{multirow}
\def\a{{\mathbf a}}
\def\b{{\mathbf b}}

\def\e{{\mathbf e}}
\def\f{{\mathbf f}}

\def\u{{\mathbf u}}

\def\x{{\mathbf x}}

\def\z{{\mathbf z}}
\def\A{{\mathbf A}}

\def\D{{\mathbf D}}

\def\L{{\mathbf L}}

\def\bR{{\mathbb{R}}}

\def\cD{\mathcal{D}}

\def\cL{\mathcal{L}}

\def\cT{\mathcal{T}}

\def\cX{\mathcal{X}}
\def\cY{\mathcal{Y}}

\newcommand{\ie}{\emph{i.e.}\@\xspace}
\newcommand{\eg}{\emph{e.g.}\@\xspace}
\newcommand{\cf}{\emph{cf.}\@\xspace}
\newcommand{\vs}{\emph{vs.}\@\xspace}

\newcommand{\apriori}{\emph{a priori}\xspace}

\renewcommand{\paragraph}[1]{\vspace{0.25\parskip}\noindent\textbf{#1}\hspace{1em}}

\newcommand{\github}{\url{https://github.com/danielegrattarola/GINR}}

\title{Generalised Implicit Neural Representations}

\author{Daniele Grattarola \\
	EPFL \\
  	Lausanne, Switzerland \\
	\texttt{daniele.grattarola@epfl.ch} \\
	\And
	Pierre Vandergheynst \\
	EPFL \\
  	Lausanne, Switzerland \\
	\texttt{pierre.vandergheynst@epfl.ch}
}

\begin{document}
\maketitle

\begin{abstract}
We consider the problem of learning implicit neural representations (INRs) for signals on non-Euclidean domains.
In the Euclidean case, INRs are trained on a discrete sampling of a signal over a regular lattice. 
Here, we assume that the continuous signal exists on some unknown topological space from which we sample a discrete graph.
In the absence of a coordinate system to identify the sampled nodes, we propose approximating their location with a spectral embedding of the graph. 
This allows us to train INRs without knowing the underlying continuous domain, which is the case for most graph signals in nature, while also making the INRs independent of any choice of coordinate system.
We show experiments with our method on various real-world signals on non-Euclidean domains.
\end{abstract}

\begin{figure}[!h]
    \centering
    \includegraphics[width=0.85\textwidth]{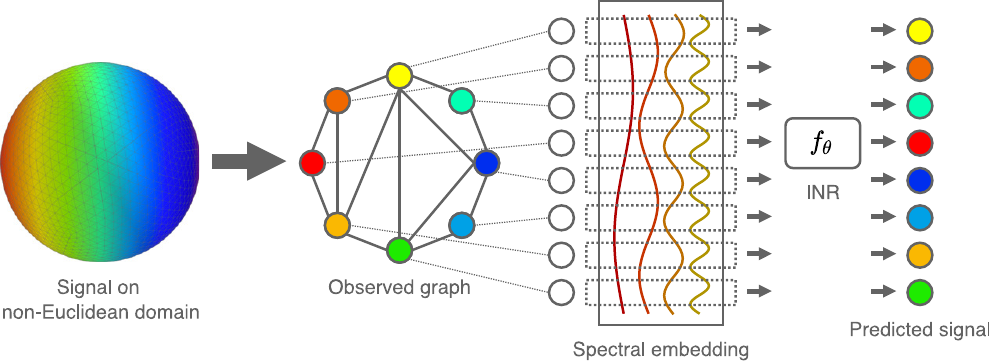}
    \caption{Given a continuous signal on a non-Euclidean domain, we observe a discrete graph realisation of it. A generalised implicit neural representation is a neural network $f_\theta$ trained to map a spectral embedding of each node to the corresponding signal value.}
    \label{fig:scheme}
\end{figure}

\section{Introduction}
\label{sec:introduction}

Implicit neural representations (INRs) are a class of techniques to parametrise signals using neural networks~\cite{stanley2007compositional,park2019deepsdf,sitzmann2020implicit,lipman2021phase}. 
INRs are trained to map each point in a given domain to the corresponding value of a signal at that point.
For example, INRs for images learn to map the 2D coordinates of pixels to their corresponding RGB values.
INRs can also be conditioned on a latent vector, typically learned end-to-end as part of the model, that allows them to represent different signals on the same domain. 
INRs have been successfully applied to model complex signals like images~\cite{stanley2007compositional,sitzmann2020implicit}, signed distance functions~\cite{park2019deepsdf}, and radiance fields~\cite{mildenhall2020nerf}.

By expressing a signal as an INR, we obtain a continuous approximation of the signal on the whole domain. This allows us to compute a higher-resolution approximation of the signal by sampling more points on the domain (\eg, a finer grid of pixels). 
Additionally, since INRs are fully differentiable, the latent vector of a conditional INR can be optimised with backpropagation and gradient descent to obtain a signal with some desired characteristics. 
For example, a conditional INR for signed distance functions can be used to design surfaces with a target aerodynamic profile~\cite{remelli2020meshsdf}. 

So far, the literature on INRs has only focused on signals on Euclidean domains.
However, signals defined on non-Euclidean domains are ubiquitous in nature and are especially relevant in artificial intelligence, as demonstrated by the recent rise of geometric deep learning~\cite{bronstein2017geometric}.
In this paper, we propose an extension of the INR setting to signals on arbitrary non-Euclidean domains. 

\paragraph{Contributions} We formulate the \emph{generalised} INR problem as the task of learning an implicit representation for a signal on an arbitrary topological space $\cT$ where, instead of observing the signal sampled on a regular lattice, we observe a graph (\ie, a discretisation of $\cT$) and the corresponding graph signal. 
In most practical cases, $\cT$ is unknown and we cannot represent the sampled vertices in a coordinate system.
Even when $\cT$ is known (as in the Euclidean case), training an INR on a fixed coordinate system means that the model will depend on this choice.
We solve both these issues by identifying the sampled nodes with an intrinsic spectral embedding obtained from the eigenvectors of the graph Laplacian.
We then train a neural network to map the spectral embeddings to the corresponding signal values. Figure~\ref{fig:scheme} shows a schematic view of the method.

Since the eigenvectors of the graph Laplacian are a discrete approximation of the continuous eigenfunctions of the Laplace-Beltrami operator on $\cT$ (when appropriately rescaled)~\cite{belkin2001laplacian,bengio2003spectral,belkin2006convergence}, at inference time we can map arbitrary points on $\cT$ to the corresponding approximation of the signal, as long as the discrete graph signal is sampled consistently. This allows us to compute higher-resolution signals or to estimate the signal on different graph realisations of the same phenomenon (\eg, different social networks with a similar structure).

\paragraph{Results} In the experiments section, we show concrete examples of learning INRs for signals on graphs and manifolds, using real-world data from biology and social networks. 
We also show that INRs trained on one instance of a graph can be transferred to different graph realisations of the underlying domain. 
Then, we show applications of conditional generalised INRs to model different signals on the same irregular domain, and also different signals on different domains. 
Finally, we conclude the paper with an experiment that consolidates all our results, modelling real-world meteorological signals on the spherical surface of the Earth.

\section{Related works}
\label{sec:related_works}

INRs have recently attracted the attention of the machine learning community for their ability to represent complex signals, especially in computer vision, 3D rendering, and image synthesis applications~\cite{park2019deepsdf,mescheder2019occupancy,atzmon2020sal,gropp2020implicit,mildenhall2020nerf,skorokhodov2021adversarial,li20223d}.
Recent work has highlighted that INRs benefit from computing sinusoidal transformations of the coordinates, typically called Fourier features or positional encodings (PEs)~\cite{vaswani2017attention,tancik2020fourier,mildenhall2020nerf,yariv2020multiview,benbarka2022seeing}. 
Since the PEs typically used in INRs are also the eigenfunctions of the Laplace operator in Euclidean domains, our method is a generalisation of this approach.

However, these transformations are not applicable to non-Euclidean domains and still depend on a choice of coordinate system.
A proposal for making INRs equivariant under the SO(3) group has been presented by \citet{deng2021vector}, although this technique does not apply in general to non-Euclidean domains. 
The field of geometric deep learning has also explored the use of graph PEs similar to the ones we use in this paper, showing their usefulness as node features for graph neural networks~\cite{srinivasan2019equivalence,loukas2019graph,dwivedi2020benchmarking,dwivedi2020generalization}. Recent research has also investigated the idea of learnable graph PEs~\cite{dwivedi2022graph,chamberlain2021beltrami}.

Related to our work, \citet{belkin2001laplacian}, \citet{boguna2021network} and \citet{levie2021transferability} have studied the geometrical link between graphs and topological spaces. \citet{keriven2020convergence,keriven2021universality} investigated the convergence, in the limit, of graph neural networks trained on large graphs. 
An alternative view of graphs as discrete samples of continuous spaces is the concept of graphon, which can be seen both as a generative model and as the limit of a sequence of graphs~\cite{lovasz2012large}.

We also mention the work of \citet{koestler2022intrinsic}, who propose a similar idea to ours for learning fields on meshes. In this work, the authors train a neural network to map the eigenfunctions of the Laplacian to an RGB texture.
By contrast, we focus on a wider range of tasks and experiments, including transferring to completely different graphs, conditioning the INRs, and modelling dynamical systems. 
Our scheme to compute generalised embeddings is also different from Koestler et al.'s, not relying on knowing node coordinates.

\section{Method}
\label{sec:method}

\paragraph{Standard setting}
In the standard INR setting, we consider a signal $f: \cX \rightarrow \cY$ with $\cX \subseteq \bR^d$ and $\cY \subseteq \bR^p$. 
We observe a discrete realisation of the signal $f(\x_i)$ for $i=1, \dots, n$, where points $\x_i$ are sampled on a regular lattice on $\cX$.
Then, we train a neural network $f_\theta: \cX \rightarrow \cY$, with parameters $\theta$, on input-output pairs $(\x_i, f(\x_i))$. Since we know that the signal domain is a subset of $\bR^d$, at inference time we can sample points anywhere on $\cX$ to compute the approximated value of the signal at those points. 
For example, an INR for images maps equispaced points in the unit square (pixel coordinates) to points in the unit cube (RGB values normalised between 0 and 1). The specific image on which we train the INR is a realisation of one such signal at a given resolution, and at inference time we can sample a finer lattice to super-resolve the image~\cite{stanley2007compositional,sitzmann2020implicit}.

\paragraph{Generalised setting}
In the generalised setting, we consider a continuous signal $f: \cT \rightarrow \cY$, with $\cT$ an arbitrary topological space. 
We observe a discrete graph signal $f(v_i)$ on an undirected graph $G = (V, E)$, with node set $V=\{v_i\}$ for $i=1,\dots,n$ and edge set $E \subseteq V \times V$. Note that the graph can be weighted if a metric is available on $\cT$.
The meaning of sampling $G$ from $\cT$ is intuitive in the case of geometric meshes or other physical structures like proteins (in which case we also know the coordinates of $v_i$), but the same reasoning also applies to more complex domains with an abstract meaning (\eg, the space of all possible papers and their citations). 
We generally assume that the graph describes some measure of closeness between uniformly sampled points on $\cT$, be it a function of the coordinates (\eg, a kernel) or some logical or functional relation that is given as part of the data.
In general, we don't assume to know the true $\cT$ from which $G$ is sampled or the coordinates of $v_i$. We only observe $G$ and the associated graph signal.
For an in-depth discussion on the relation between graphs and topological spaces, see references \cite{belkin2001laplacian}, \cite{boguna2021network}, and \cite{levie2021transferability}.

\paragraph{Proposed method}
We approach the generalised INR problem by mapping $v_i$ to $f(v_i)$ through a spectral embedding obtained from the graph Laplacian. 
Let $\A \in \bR^{n \times n}$ be the weighted adjacency matrix of graph $G$, $\D$ the diagonal degree matrix, and $\L = \D - \A$ the combinatorial Laplacian. 
The eigendecomposition of the Laplacian yields an orthonormal basis of eigenvectors $\{\u_k \in \bR^n\}$ for $k=1, \dots, n$, with a canonical ordering given by their associated eigenvalues $\lambda_1 \leq \lambda_2 \leq \dots \leq \lambda_n$. 
We define a generalised spectral embedding of size $k$ for node $v_i$ as 
\begin{equation}
\e_i = \sqrt{n} \left[\u_{1, i}, \dots, \u_{k, i}\right]^\top \in \bR^k.
\end{equation}
\begin{wrapfigure}{r}{0.45\textwidth}
    \vspace{-0.5cm}
    \centering
    \includegraphics[width=0.45\textwidth]{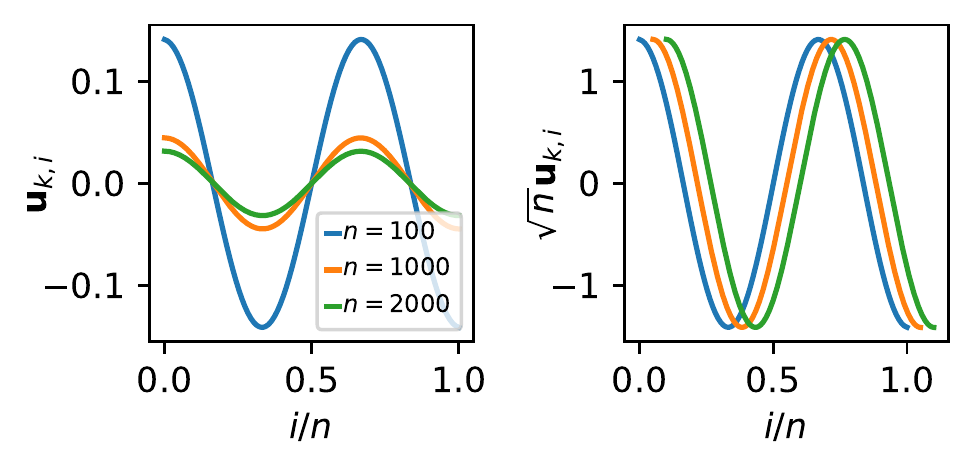}
    \caption{Laplacian eigenvectors (left) \vs rescaled spectral embeddings (right) of a path graph, for ${k=3}$ and ${n=100, 1000, 2000}$. The curves on the right are shifted to improve visualisation.}
    \vspace{-0.5cm}
    \label{fig:comparison_eigv_embs_path_graph}
\end{wrapfigure}
The generalised spectral embeddings are a discrete approximation of the eigenfunctions of the continuous Laplace-Beltrami operator on $\cT$ and converge to them for $n \rightarrow \infty$.
See the discussion in references~\cite{belkin2001laplacian}, \cite{bengio2003spectral}, \cite{bengio2003out}, and~\cite{belkin2006convergence} for a formal analysis.
In practice, rescaling the eigenvectors by $\sqrt{n}$ ensures that the embeddings for graphs of different sizes will have the same range component-wise and that similar nodes will have similar embeddings regardless of the graph size. As an example, we show a comparison between the eigenvectors and the generalised spectral embeddings of a path graph in Figure~\ref{fig:comparison_eigv_embs_path_graph}.

We train a neural network $f_\theta: \e_{i} \mapsto f(v_i)$ on the observed graph signal. 
Since the generalised spectral embeddings are an intrinsic property of the graph, the INR will not depend on any choice of coordinate system but only on the topology of the underlying continuous domain. 

At inference time, we can compute the approximated value of the signal at an arbitrary location on $\cT$ by computing the associated spectral embedding.
If we know $\cT$, or if we estimate it, we can sample new vertices directly from the domain and apply the same procedure used to construct the training graph. If $\cT$ is unknown (\eg, in the case of natural graphs that describe some abstract concept like citations, social interactions, or biological relations), then we just assume to observe a similar graph sampled from $\cT$.
An important difference between generalised INRs and Euclidean INRs is that generally we must observe the full graph in order to compute the inputs. This is necessary because, while in the Euclidean case we completely know the domain of the signal \apriori, in the generalised case we need to estimate the topology of the domain by sampling.

\paragraph{Complexity} Although computing the full eigendecomposition of the Laplacian has a complexity of $O(n^3)$, here we are only interested in the first $k$ eigenvectors. Also, the Laplacian of most graphs is very sparse, with nodes having an average degree $\bar d \ll n$.
We can therefore use the implicitly restarted Lanczos method for eigendecomposition implemented by the ARPACK software, which has a complexity of $O(\bar d n^2)$ and can be easily parallelised~\cite{lehoucq1998arpack}. In practice, all computations for this paper were easily managed on a commercial laptop with 10 CPU cores, scaling up to graphs in the order of $10^5$ nodes.

We also note that at inference time, depending on how the edges are constructed, it could be possible to add new nodes and edges to the training graph instead of sampling an entirely new graph from $\cT$. In this case, the spectral embeddings for the new nodes can be estimated using the Nystr\"om method without needing to compute the full eigendecomposition~\cite{baker1977numerical,bengio2003out}.

\paragraph{Alternative approaches} The generalised Laplacian embeddings used in this paper are only one of many possibilities to represent nodes sampled from $\cT$. To name a few, locally linear embeddings~\cite{roweis2000nonlinear}, Isomap~\cite{tenenbaum2000global}, Laplacian eigenmaps~\cite{belkin2003laplacian} and diffusion maps~\cite{coifman2005geometric} are all based on the idea of embedding data using the first few principal eigenvectors of a similarity matrix. Here we focus on the Laplacian since it is well known, sparse, and easy to compute, and its eigendecomposition is stable to graph perturbations. One disadvantage of Laplacian eigenvectors is that they are only unique up to sign, leading to $2^k$ possible eigenbases for a given $\cT$ and the consequent ambiguity when transferring an INR to a different graph realisation.
Additionally, eigenvalues of multiplicity greater than 1 also introduce ambiguity in the eigenvectors, since all rotations and reflections of the associated eigenspace are valid choices. 
However, simple heuristics can be used to eliminate sign ambiguity, and we did not encounter any other practical issue in this regard. Alternative ways to resolve the ambiguity is to use PEs based on random walks~\cite{dwivedi2022graph,mialon2021graphit,li2020distance}, heat kernels~\cite{sun2009concise,feldman2022weisfeiler}, or the more recent sign-and-basis invariant neural networks~\cite{lim2022sign}. We leave the exploration of these alternatives to future work, since they do not significantly impact our main contributions.

\section{Experiments}
\label{sec:experiments}

\paragraph{Setting} We implement the generalised INR as a SIREN multi-layer perceptron~\cite{sitzmann2020implicit}. The model has 6 layers with 512 hidden neurons and a skip connection from the input to the middle layer. We use the same hyperparameters and initialisation scheme suggested by \citet{sitzmann2020implicit}. We train the model using Adam~\cite{kingma2014adam} with a learning rate of $10^{-4}$ and an annealing schedule that halves the learning rate if the loss does not improve for 1000 steps. At each step, we sample 5000 nodes randomly from the graph as a mini-batch. We use spectral embeddings of size $k=100$, a value that we found empirically as a good trade-off between complexity and performance. We report an analysis on the effect of $k$ in Section~\ref{sec:learning_generalised_INRs}.
We ran all experiments on an Nvidia Tesla V100 GPU.
The code to reproduce our results and the high-resolution version of all figures are available at \github.
All deviations from the default setting are documented in the supplementary material, where we also report an additional experiment on solving differential equations with generalised INRs.

\subsection{Learning generalised INRs}
\label{sec:learning_generalised_INRs}

We begin by verifying that a typical INR architecture can successfully learn arbitrary graph signals in the generalised setting. 
We consider three datasets as test cases.

\begin{figure}
    \centering
    \includegraphics[width=0.2\textwidth]{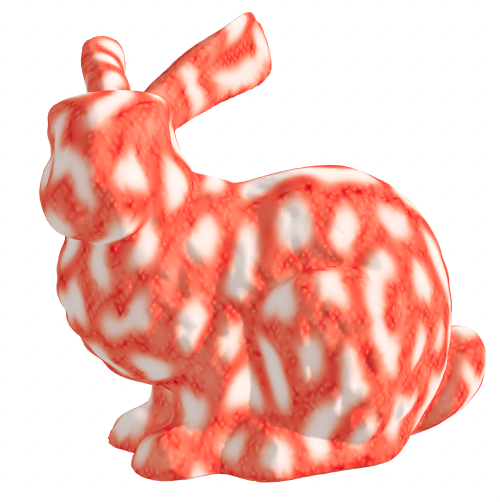}
    \includegraphics[width=0.2\textwidth]{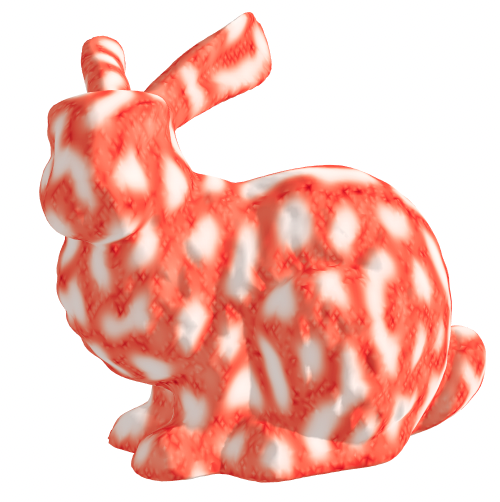}
    \includegraphics[width=0.2\textwidth]{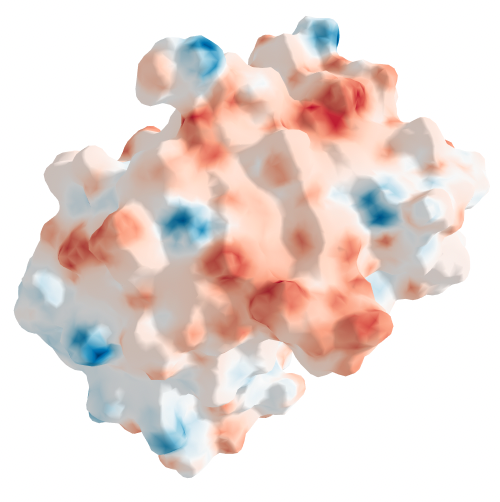}
    \includegraphics[width=0.2\textwidth]{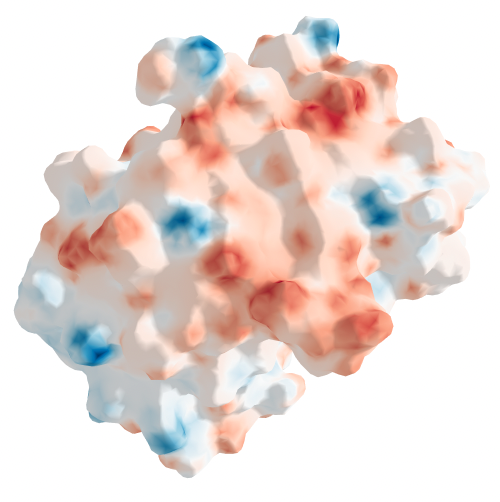}
    \caption{Ground truth signals \vs signals predicted by the INR. Best viewed in colour.}
    \label{fig:inr_comparison_true_v_pred}
\end{figure}

\paragraph{Bunny} We generate a texture on the Stanford bunny mesh\footnote{Available at \url{https://graphics.stanford.edu/data/3Dscanrep/}} using the Gray-Scott reaction-diffusion model given by:
\begin{equation}
\Delta\a = - D_a\L\a - \a\odot\b\odot\b + F(\mathbf{1}_n-\a); \;\; 
\Delta\b = - D_b\L\b + \a\odot\b\odot\b - (F + K)\b
\label{eq:gray-scott}
\end{equation}
for state vectors $\a, \b \in \bR^n$ and parameters $D_a, D_b, F, K \in \bR$, where $\odot$ indicates element-wise product~\cite{gray1984autocatalytic}. 
The mesh has 34834 nodes and 104288 edges.
We configure the system to yield ``coral'' patterns\footnote{$D_a=0.64, D_b=0.32, F=0.06, K=0.062$} and evolve the simulation for $10^4$ steps starting from random positive values for $\a, \b$. At the end of the evolution, we use vector $\a$ as target graph signal.
\begin{wraptable}{r}{0.5\textwidth}
    \centering
    \caption{$R^2$ and mean squared error for the bunny, protein, and US election signals.}
    \resizebox{0.5\textwidth}{!}{%
    \begin{tabular}{@{}llll@{}}
        \toprule
        & \textbf{Bunny} & \textbf{Protein} & \textbf{US Election} \\ \midrule
        $R^2$ & 1.000 & 1.000 & 0.999 \\
        MSE & $9.14 \cdot 10^{-8}$ & $1.17 \cdot 10^{-10}$ & $1.45 \cdot 10^{-3}$ \\
        \bottomrule
    \end{tabular}
    }
    \label{tab:learning_ginr_results}
    \vspace{-2\baselineskip}
\end{wraptable}
\begin{wraptable}{r}{0.5\textwidth}
    \vspace{-1.25\baselineskip}
    \centering
    \caption{Performance comparison ($R^2$) of typical INRs and generalised INRs (GINRs), using ReLU (r) or SIREN (s) activation.}
    \resizebox{0.5\textwidth}{!}{%
    \begin{tabular}{@{}lllll@{}}
        \toprule
        & \textbf{INR (r)} & \textbf{INR (s)} & \textbf{GINR (r)} & \textbf{GINR (s)} \\ \midrule
        Bunny & 0.000 & 0.919 & \textbf{0.932} & 0.885\\
        Protein & 0.799 & 0.275 & \textbf{0.921} & 0.916 \\
        \bottomrule
    \end{tabular}
    }
    \label{tab:learning_ginr_results_baseline}
    \vspace{-1\baselineskip}
\end{wraptable}

\paragraph{Protein} As a real-world domain, we consider the solvent excluded surface of a protein structure.\footnote{Protein Data Bank identifier: 1AA7}
The continuous signal is the value of the electrostatic field generated by the amino acid residues at the surface.
Protein function can be largely understood by studying their surfaces and the corresponding chemical features, like the electrostatic charge, so this represents a potentially interesting application of INRs to biology. 
We use a combination of the MSMS~\cite{sanner1996reduced} and APBS~\cite{jurrus2018improvements} software packages to estimate the surface, sample the nodes, and get the corresponding signal, following the same workflow of~\citet{gainza2020deciphering}. The final graph has 11966 nodes and 35892 edges. 

\paragraph{US election} Finally, we consider a social network dataset introduced by \citet{jia2022unifying}, in which nodes represent United States counties and edges are estimated using the Facebook Social Connectedness Index. The target signal represents the county-wise outcome of the 2012 US presidential election, as values in the range $[-1, 1]$ (indicating the proportion of votes in favour of one candidate \vs the other). The graph has 3106 nodes and 22574 edges.

\paragraph{Results} In all cases, the generalised INR achieves an $R^2$ close to 1 indicating that the model can indeed learn non-trivial signals on non-Euclidean domains. 
We report the $R^2$ and mean squared error for all datasets in Tab.~\ref{tab:learning_ginr_results}.
We compare the true graph signals and those learned by the INR, for the bunny and protein, in Fig.~\ref{fig:inr_comparison_true_v_pred}. 
As a second test, we compare typical INRs trained on node coordinates with generalised INRs trained on the spectral embeddings. 
We consider the bunny and protein since they have input coordinates for the typical INRs.
We also test different activation functions---ReLU or SIREN.
We report in Tab.~\ref{tab:learning_ginr_results_baseline} the $R^2$ for a held-out set of nodes (to evaluate whether the INRs are overfitting instead of learning a meaningful representation).\footnote{We report training details in the supplementary material.}
All results are averaged over 5 runs and have negligible standard deviations.
We see that the generalised INRs perform almost always better than the typical ones.
Note that these experiments only verify that the model can indeed learn the target signals. We will test the transferability of the model in-depth in Sec.~\ref{sec:transferability_of_generalised_INRs}.
In the following sections, we mostly focus on known spaces $\cT$ or synthetic graphs so that we can better answer other research questions by having full control of the data. 

\begin{figure}
    \centering
    \begin{subfigure}[T]{0.3\textwidth}
        \includegraphics[width=\textwidth]{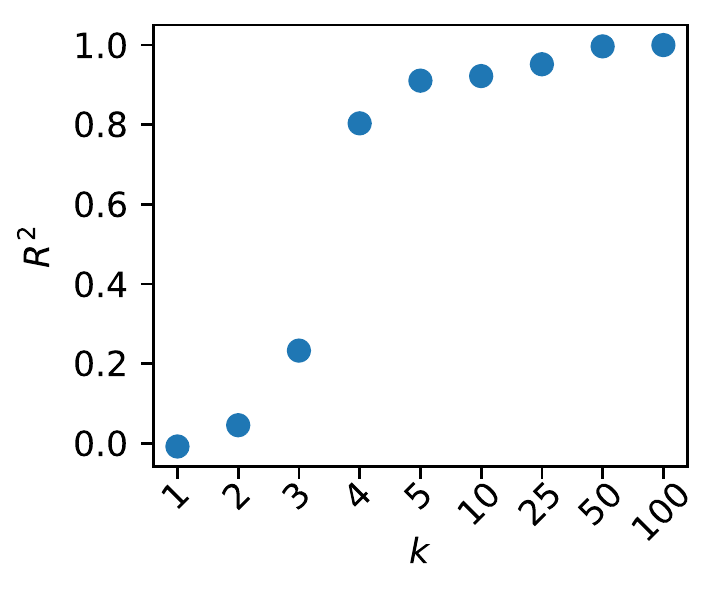}
    \end{subfigure}
    \begin{subfigure}[T]{0.2\textwidth}
        \includegraphics[width=\textwidth]{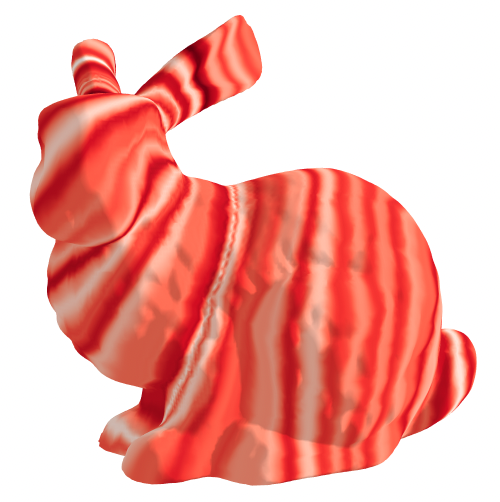}
        \caption{$k=2$}
    \end{subfigure}
    \begin{subfigure}[T]{0.2\textwidth}
        \includegraphics[width=\textwidth]{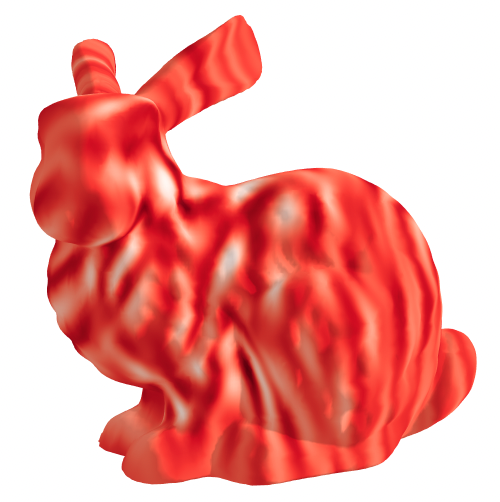}
        \caption{$k=3$}
    \end{subfigure}
    \begin{subfigure}[T]{0.2\textwidth}
        \includegraphics[width=\textwidth]{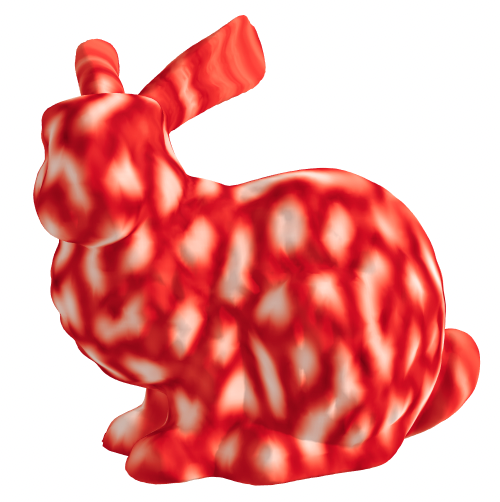}
        \caption{$k=4$}
    \end{subfigure}
    \caption{\textbf{Left:} $R^2$ \vs $k$; \textbf{Right:} signals learned by the INR for $k=2, 3, 4$.}
    \label{fig:k_ablation_results}
\end{figure}

\paragraph{Size of the spectral embeddings}
Typical INRs take as input the $d$-dimensional coordinates of points on the sampled lattice. 
In our general setting, the dimension $k$ of the spectral embeddings is a hyperparameter of the method. 
To evaluate the impact of $k$ on the performance of the INR, we train the same model for different values of $k$ on the Stanford bunny. 

\begin{wrapfigure}{r}{0.35\textwidth}
    \vspace{-0.4cm}
    \centering
    \includegraphics[width=0.3\textwidth]{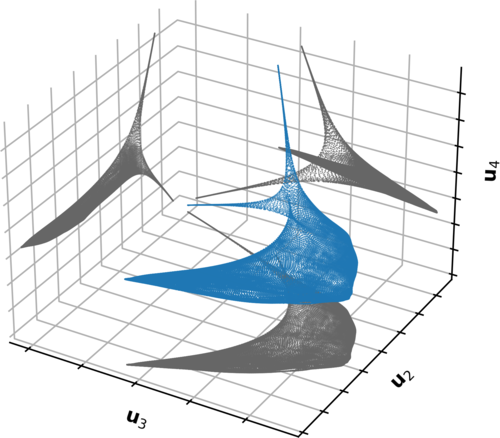}
    \caption{Spectral drawing (blue) of the bunny using the first three non-trivial eigenvectors. In grey, 2d projections of the 3d drawing.}
    \label{fig:bunny_spectral_embedding}
    \vspace{-3\baselineskip}
\end{wrapfigure}

We report the results in Figure~\ref{fig:k_ablation_results}, showing that the model fails to learn a meaningful representation for $k \leq 3$.
This is reasonable because at least 3 non-trivial eigenvectors are needed to correctly distinguish the main structural features of the bunny (ears, tail, etc.), as shown in Figure~\ref{fig:bunny_spectral_embedding}. 
We also see that, while the INR has no issues in learning the signal for $k=100$, for lower values of $k$ the model struggles to represent the signal on the ears. 
Figure~\ref{fig:bunny_spectral_embedding} again provides a possible explanation for this result, since the first few eigenvectors tend to collapse on the ears and higher frequency eigenvectors are needed to distinguish points towards the narrow tips. We will show another consequence of this in Section~\ref{sec:transferability_of_generalised_INRs}. We also mention that the drop in performance for lower values of $k$ on the other two datasets is less pronounced than on the bunny, likely because of their lack of any high-frequency features.

\subsection{Transferability of generalised INRs}
\label{sec:transferability_of_generalised_INRs}

Because the generalised spectral embeddings are a discrete approximation of the continuous eigenfunctions of the Laplace-Beltrami operator on $\cT$, we can apply a trained INR to a different graph realisation of $\cT$ as long as its spectral structure is consistent with the training data. 

In practical terms, we need to know 1) how to sample new nodes and 2) how the nodes are connected. If we know $\cT$, then the problem simply boils down to choosing a sampling strategy for the points and a suitable measure of closeness (\eg a kernel or edge-generating function like the k-nearest neighbours algorithm).
If $\cT$ is unknown, then we assume to observe a new graph through the same process that generated the training data (\eg, we observe new social interactions among the same or a similar set of people).

Note that, when sampling a new graph from $\cT$, the spectral embeddings will be a slightly different (possibly better) approximation of the continuous eigenfunctions. As this effect is more evident on high-frequency eigenvectors, we empirically observed that training the INR with a smaller $k$ and ReLU activations improved its transferability (\cf Table~\ref{tab:learning_ginr_results_baseline}). See the supplementary material for more details.
We investigate the transferability of generalised INRs in two settings.

\paragraph{Transferring to similar graphs} As a first example, we study the case in which $\cT$ is unknown and we only observe the discrete graphs. We consider a stochastic block model (SBM) with two communities of equal size (in total, $n=1000$), parametrised by an inter-connection probability $p$ and an intra-connection probability $r$.

We train an INR on a graph with a strong community structure ($p=0.1, r=0.5$) and a graph signal indicating the community of each node (effectively a node classification task). 
Then, we test the INR on graphs sampled in the ranges $p \in [0.1, 1]$ and $r \in [0.1, 1]$ to test the robustness of the model to changes in the community structure. 

\begin{figure}
    \centering
    \begin{subfigure}[T]{0.35\textwidth}
        \includegraphics[width=\textwidth]{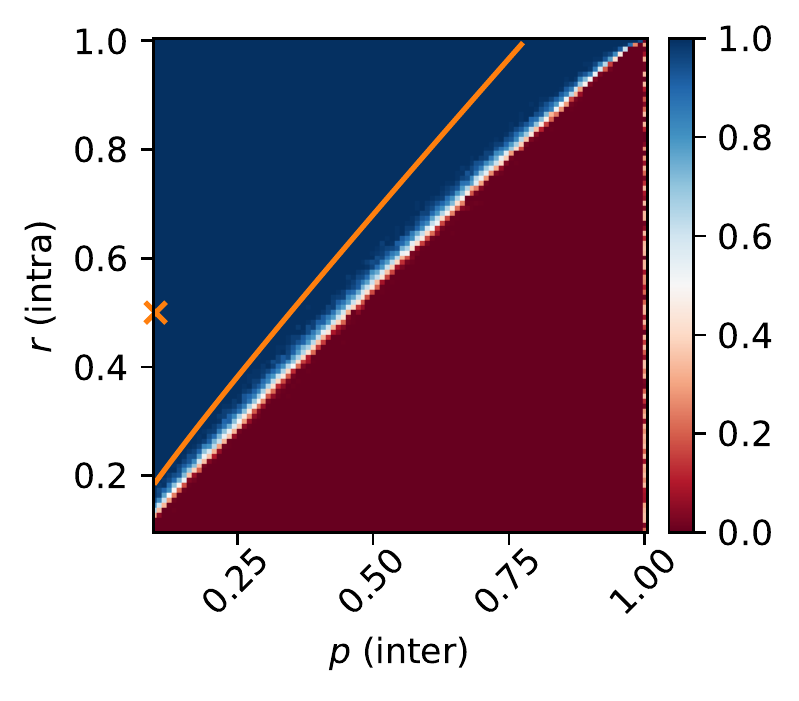}
    \end{subfigure}
    \begin{subfigure}[T]{0.25\textwidth}
        \includegraphics[width=\textwidth]{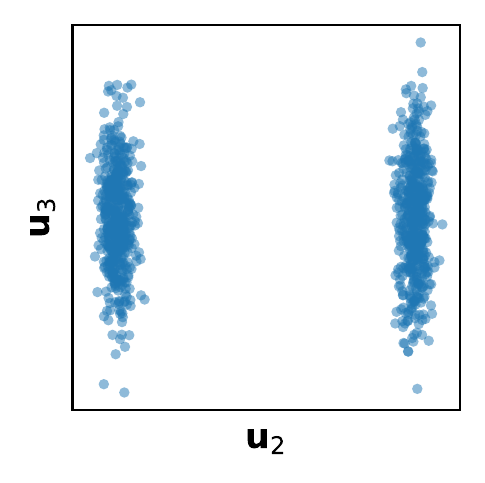}
        \caption{$p=0.1, r=0.5$}
    \end{subfigure}
    \begin{subfigure}[T]{0.25\textwidth}
        \includegraphics[width=\textwidth]{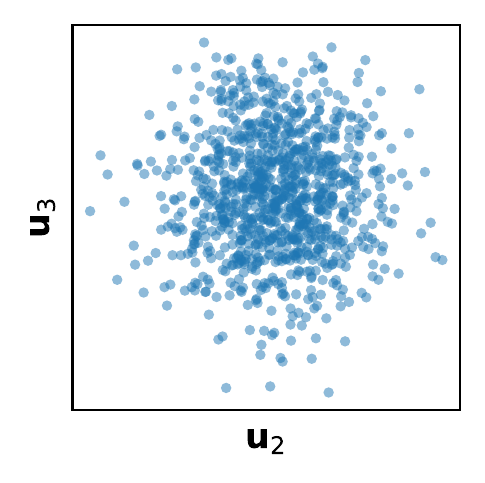}
        \caption{$p=0.5, r=0.1$}
    \end{subfigure}
    \caption{\textbf{Left:} normalised mutual information at test time, evaluating on graphs with different inter- and intra-connection probabilities. We trained the INR on $p=0.1$ and $r=0.5$, marked with $\times$ on the plot. The orange line indicates the theoretical boundary above which it is possible to perfectly recover the true signal. Best viewed in colour. \textbf{Right:} the first two non-trivial eigenvectors of two SBMs, respectively exhibiting strong and no community structure.}
    \label{fig:sbm_results}
\end{figure}

We report our results in Figure~\ref{fig:sbm_results}.
First, we note that perfectly recovering the correct clusters is only theoretically guaranteed in certain regions of the parameter space~\cite{abbe2017community}, delimited with an orange line in the figure.\footnote{The region is above curve $r = \ln(n)/n \cdot \left(\sqrt{2} + \sqrt{np/\ln(n)}\right)^2$~\cite{abbe2017community}.}
We see that the trained INR transfers perfectly to all graphs sampled in this region and that its performance sharply drops only for $p > r$, where graphs have no community structure (see Figure~\ref{fig:sbm_results}~a,~b).
Additionally, we see that the INR maintains a good performance also in the region between the theoretical boundary and the line $p = r$, \ie, a region in which graphs still exhibit some community structure although perfect recovery is not guaranteed. Overall, this indicates that training the INR on an instance of a graph allows us to transfer it to graphs with a similar structure.

\begin{figure}
    \centering
    \begin{minipage}{.65\textwidth}
    \includegraphics[width=0.45\textwidth]{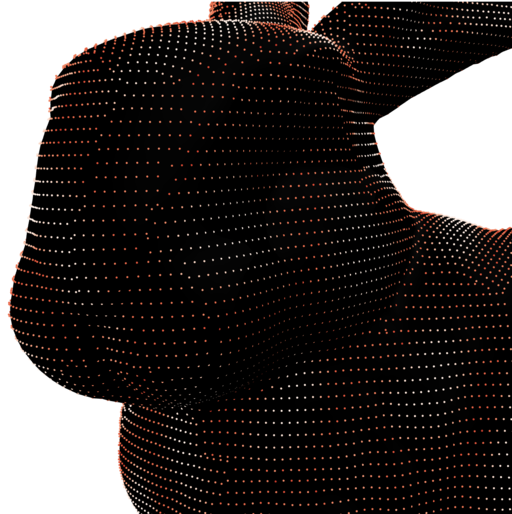}\hspace{0.5cm}
    \includegraphics[width=0.45\textwidth]{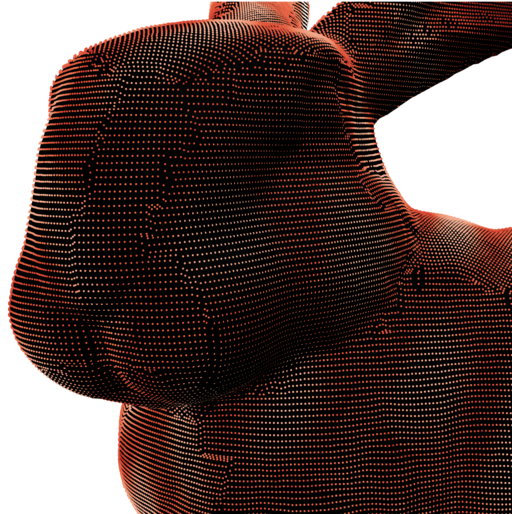}
    \caption{Zoomed-in comparison between the nodes of the training graph (left) and its super-resolved version (right), plotted on a black surface to aid visualisation. An interactive version in vector graphics of this figure is available in the supplementary material.}
    \label{fig:bunny_zoomed_in}
    \vspace{-0.5cm}
\end{minipage}\hfill
\begin{minipage}{0.3\textwidth}
    \includegraphics[width=\textwidth]{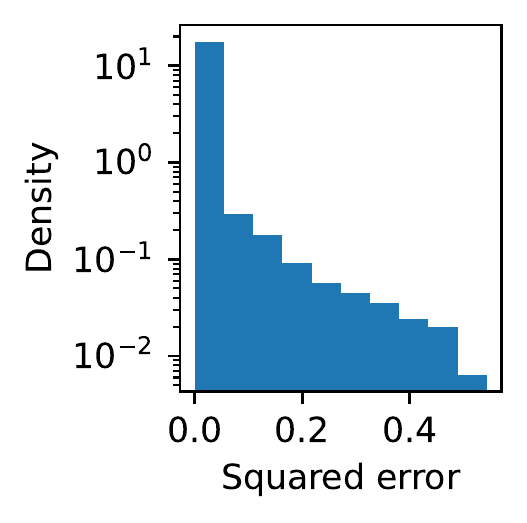}
    \caption{Distribution of the error in the super-resolution experiment.}
    \label{fig:super_resolution_error_density}
    \vspace{-0.5cm}
\end{minipage}
\end{figure}

\paragraph{Super-resolution} As a second test case, we consider the super-resolution of a signal. We consider again the Stanford bunny mesh with the same texture graph signal as Section~\ref{sec:learning_generalised_INRs} and we apply Loop's method for mesh subdivision to obtain a higher-resolution mesh with 139122 nodes and 416929 edges~\cite{loop1987smooth}.
Then, we compute the spectral embeddings for the new graph and predict the signal with the INR trained on the low-resolution graph.

Because the unit-norm eigenvectors are only unique up to sign, the INR will transfer to a different graph only if its eigenvectors are aligned to the training ones (\ie, similar nodes must have similar spectral embeddings --- in the previous experiment, we did not have to deal with this problem due to symmetry in the SBM).  However, in many cases, a simple comparison of the eigenvectors' histograms provides a useful heuristic for automatically aligning them. For simplicity, here we also verify the alignment of the two eigenbases manually.

We show in Figure~\ref{fig:bunny_zoomed_in} a qualitative comparison between the training and super-resolved signals, from which we see that the INR can predict the graph signal correctly. 
Quantitatively, since mesh subdivision does not remove the original nodes from the mesh, we can compute the $R^2$ between the training signal and part of the super-resolved signal. 
The model achieves an $R^2$ of 0.39 compared to a training $R^2$ of 0.99. However, this apparent drop in performance is mostly due to the high-frequency ears of the bunny, for which the model evidently overfits and cannot tolerate even small changes in the eigenvectors.
Figure~\ref{fig:super_resolution_error_density} shows the distribution of the squared error between the ground truth and predicted signal, from which we see that the error is close to zero for most of the nodes (note that the y-axis is logarithmic). By ignoring the nodes above the 90th percentile of the squared error distribution, the $R^2$ score is 0.94. Increasing $k$ resulted in a better prediction of the ears region (\cf Figure~\ref{fig:k_ablation_results}) at the cost of worse overall transferability. 

\subsection{Conditional generalised INRs}
\label{sec:conditional_generalised_INRs}

We also consider the setting in which a generalised INR is conditioned on a latent vector $\z \in \bR^q$ to allow the representation of different signals by the same neural network $f_\theta(\e_i, \z)$. Unlike the Euclidean setting, here we can investigate two cases: 1) learning a conditional INR for different signals on the same domain (the typical setting), and 2) learning a conditional INR for different signals on different domains. 

\paragraph{Reaction-diffusion process} For the first case, we train a generalised INR to approximate the Gray-Scott reaction-diffusion process that we used to generate the graph signal in Section~\ref{sec:learning_generalised_INRs}, given by Equation~\eqref{eq:gray-scott}. We consider a signal $f(v_i, t)$, where $t \in \mathbb{N}$ is the number of updates since the start of the simulation. 
We evolve the system for 3000 steps starting from the same random initialisation used in Section~\ref{sec:learning_generalised_INRs} and we train the INR $f_\theta(\e_i, t)$ using 300 time steps sampled at $t \equiv 0\ (\mathrm{mod}\ 10)$. 
The graph is fixed throughout the evolution, so the INR must learn to exploit the conditioning input $t$ to correctly predict the signal at different times.

\begin{figure}
    \centering
    \begin{subfigure}[T]{0.18\textwidth}
        \includegraphics[width=\textwidth]{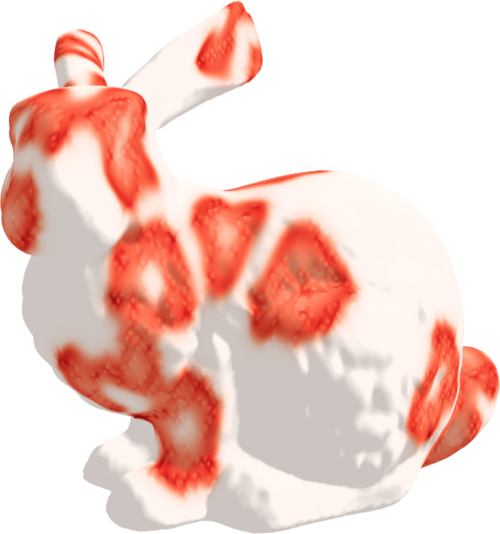}
        \caption{$t=625$}
    \end{subfigure}
    \begin{subfigure}[T]{0.18\textwidth}
        \includegraphics[width=\textwidth]{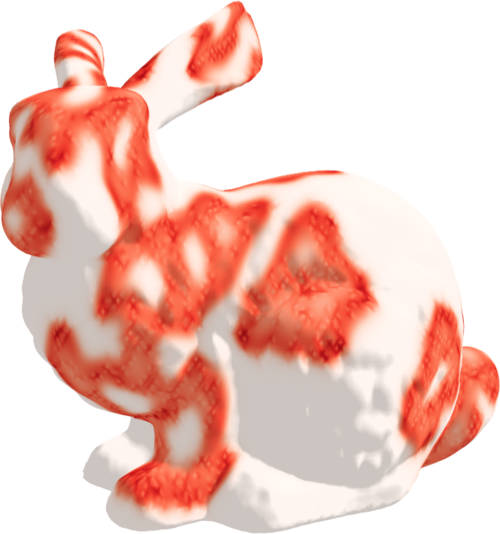}
        \caption{$t=875$}
    \end{subfigure}
    \begin{subfigure}[T]{0.18\textwidth}
        \includegraphics[width=\textwidth]{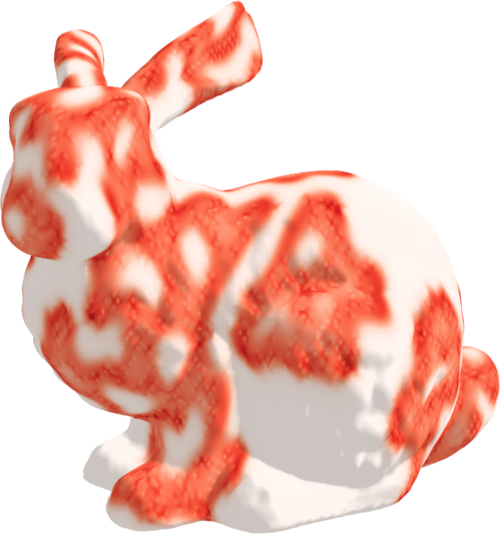}
        \caption{$t=1125$}
    \end{subfigure}
    \begin{subfigure}[T]{0.18\textwidth}
        \includegraphics[width=\textwidth]{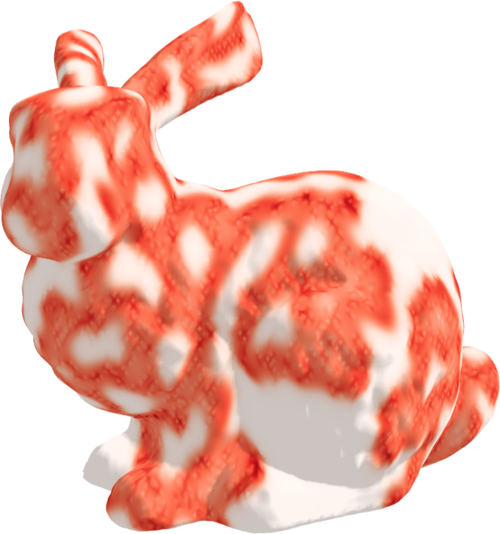}
        \caption{$t=1375$}
    \end{subfigure}
    \begin{subfigure}[T]{0.18\textwidth}
        \includegraphics[width=\textwidth]{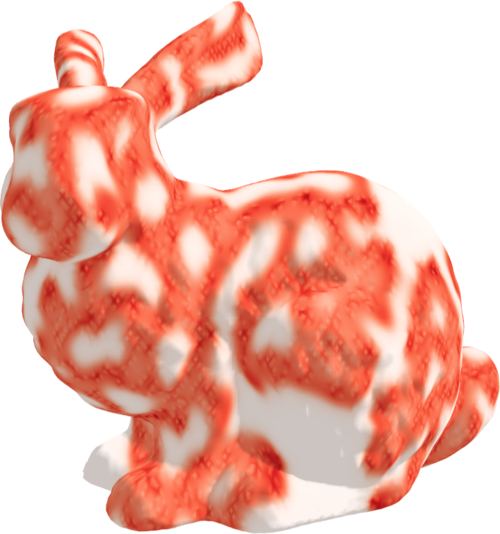}
        \caption{$t=1625$}
    \end{subfigure}
    \caption{Signals predicted by the conditional INR $f_\theta(\e_i, t)$ at equispaced time steps $t \in [625, 1625]$. All time steps shown were not part of the training set. Also, each test time step is as far as possible from its two closest training samples (\ie, training samples are at $t \equiv 0\ (\mathrm{mod}\ 10)$ while test samples are at $t \equiv 5\ (\mathrm{mod}\ 10)$). An animated version of this figure with 600 predicted samples is available in the supplementary material.}
    \label{fig:bunny_time_results}
\end{figure}

\begin{wrapfigure}{r}{0.3\textwidth}
    \vspace{-0.55cm}
    \includegraphics[width=0.3\textwidth]{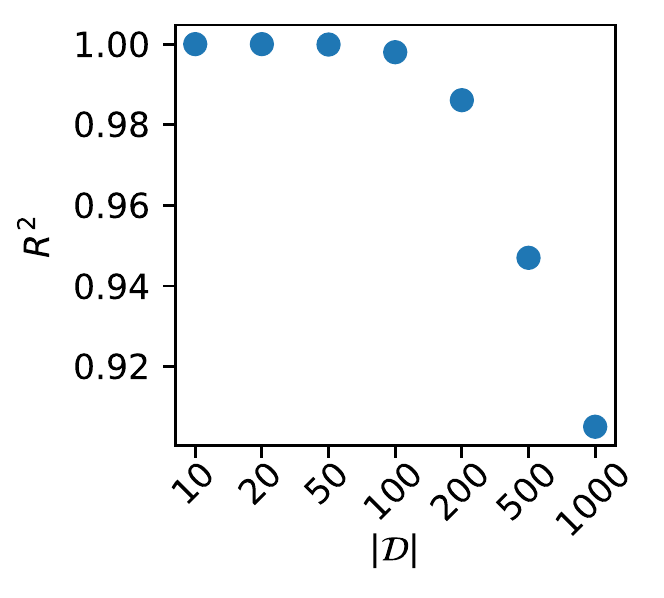}
    \caption{$R^2$ \vs size of the dataset $|\cD|$.}
    \label{fig:proteins_multi_results}
    \vspace{-0.3cm}
\end{wrapfigure}

At test time, we predict the signal for all the remaining time steps and compare the prediction with the ground truth signals. The INR achieves a test $R^2$ of $0.999 \pm 0.006$ indicating that it successfully learned to approximate the true dynamics. We report in Figure~\ref{fig:bunny_time_results} a few samples predicted by the INR at test time. 
By comparison, a simple baseline using linear interpolation between training samples achieves a similar $R^2$, while requiring to keep all 300 training steps in memory. The INR, on the other hand, represents the signal and its evolution with a single set of weights and a scalar control parameter.

\paragraph{Multi-protein INR} For the second case, we consider a dataset $\cD = \{(\cT_m, f_m)\}$ of protein surfaces $\cT_m$ with associated electrostatic signals $f_m: \cT_m \rightarrow \bR$ for $m=1, \dots, |\cD|$, obtained following the same protocol as Section~\ref{sec:learning_generalised_INRs}.
We configure the neural network $f_\theta(\e_i, \z_m)$ as an autodecoder~\cite{park2019deepsdf} where $\z_m \in \bR^z$ is a vector of free parameters learned end-to-end alongside the weights of the neural network. We assign a separate vector $\z_m$ to each training sample, allowing the model to find the best latent representation for the dataset. Since the continuous surface and the associated signals are unique to each protein in the dataset, the latent vector $\z_m$ must capture both aspects. We set the latent vector size $z=8$, which we empirically found to give consistently good performance across datasets. Larger values of $z$ did not significantly improve the performance. 

We test the ability of the model to represent datasets of different sizes, by sampling random proteins from the dataset used in reference \cite{gainza2020deciphering}. Figure~\ref{fig:proteins_multi_results} shows the $R^2$ of the model for different values of $|\cD|$. Understandably, the performance decreases as the number of samples grows, although the model achieves $R^2 \ge 0.9$ for datasets of up to 500 different proteins.
Overall, these results indicate that the model can represent many domains and signals with a single set of weights.

\subsection{Weather modelling}
\label{sec:weather_modelling}

\begin{figure}
    \centering
    \begin{subfigure}[T]{0.32\textwidth}
        \includegraphics[width=\textwidth]{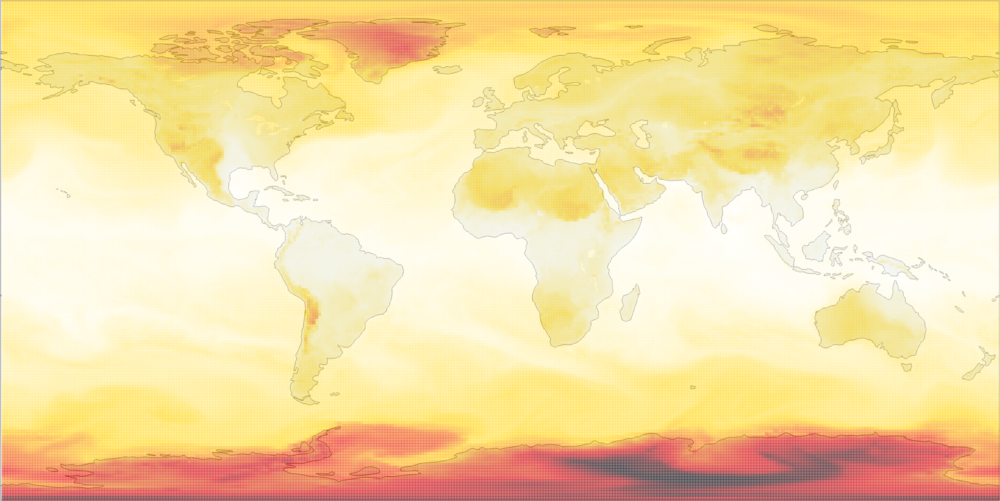}
        
        \vspace{0.1cm}
        \includegraphics[width=\textwidth]{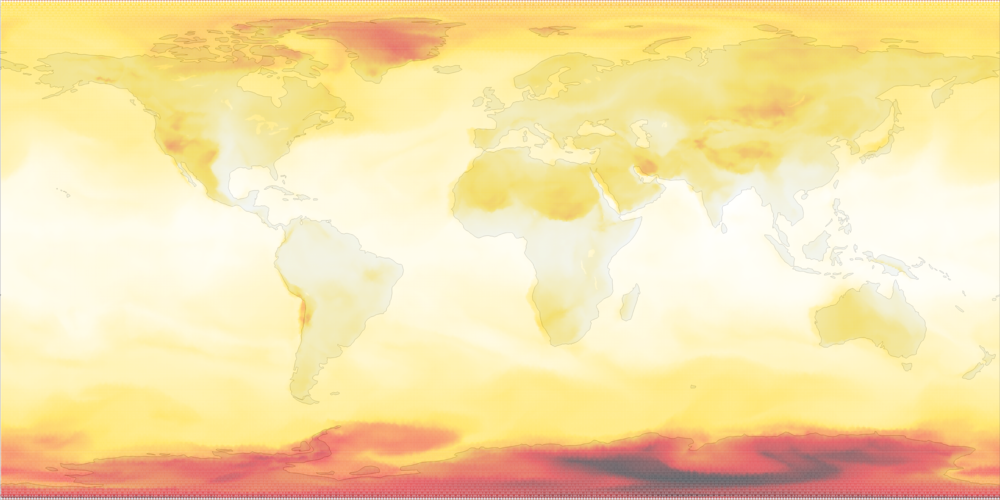}
        \caption{Temperature}
    \end{subfigure}
    \begin{subfigure}[T]{0.32\textwidth}
        \includegraphics[width=\textwidth]{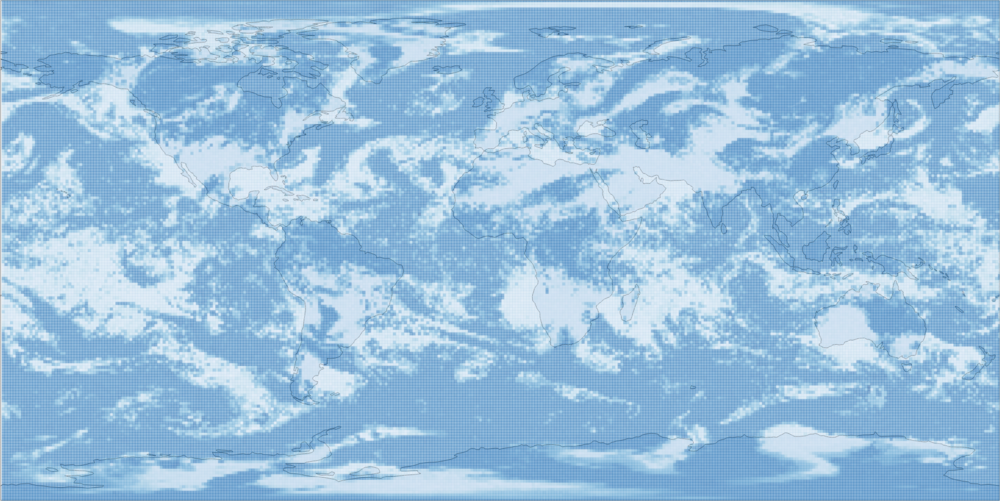}
        
        \vspace{0.1cm}
        \includegraphics[width=\textwidth]{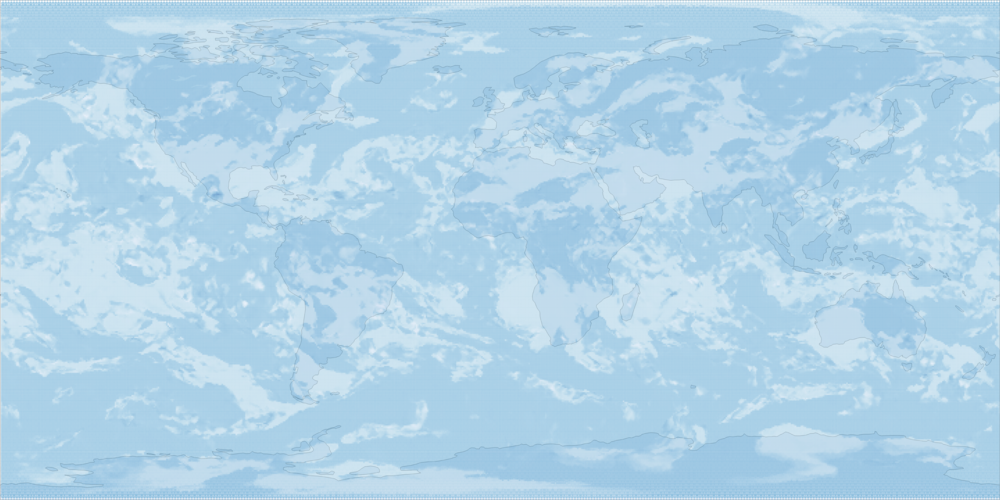}
        \caption{Clouds}
    \end{subfigure}
    \begin{subfigure}[T]{0.32\textwidth}
        \includegraphics[width=\textwidth]{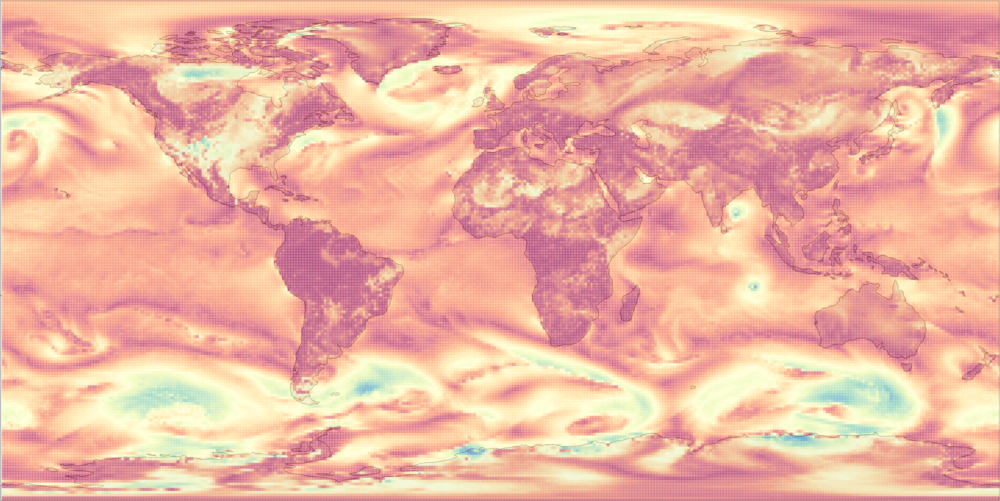}
        
        \vspace{0.1cm}
        \includegraphics[width=\textwidth]{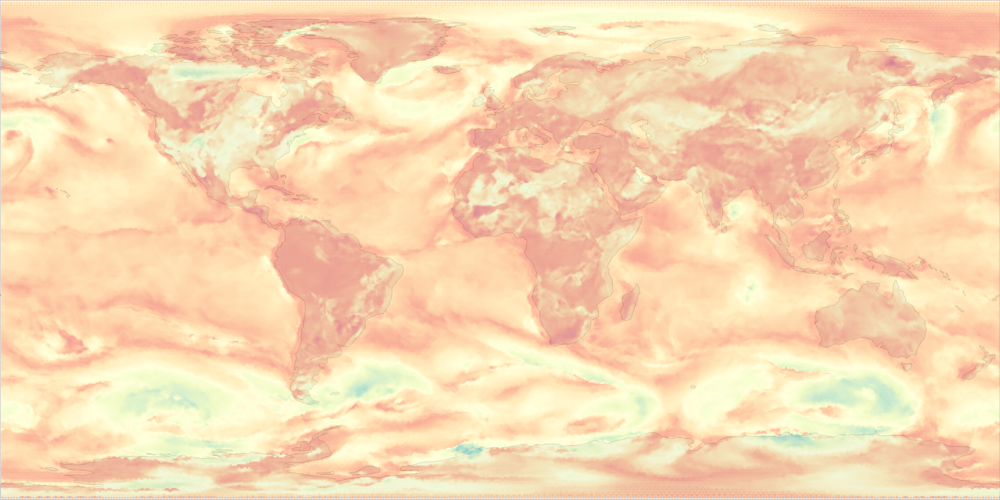}
        \caption{Wind}
    \end{subfigure}
    \caption{\textbf{Top}: prediction of the INR for $t=4.5$ (randomly chosen, not in the training data) at the original resolution. \textbf{Bottom}: The super-resolved versions of the top row. Note that the apparently lighter colours are simply artifacts of the plotting software. Best viewed in colour and high resolution.
    An animated and uncompressed version of this figure is available in the supplementary material.}
    \label{fig:weather_results}
\end{figure}
    
As a final experiment, we consolidate all results from previous sections to showcase an application of generalised INRs in a real-world setting.
We consider signals $f(v_i, t)$ representing hourly meteorological measurements on the spherical domain of the Earth's surface.
Our goal is to train a conditional INR at a given spatial and temporal resolution and then use it to super-resolve the signal over space and time.

\paragraph{Data} We collected data from the National Oceanic and Atmospheric Administration (NOAA) Operational Model Archive and Distribution System, specifically from the Global Forecast System (GFS). 
The data consists of three different signals representing: the dew point temperature 2m above-ground measured in K (\emph{dpt2m}), the atmosphere's total cloud cover percentage (\emph{tcdcclm}), and the surface wind speed in m/s (\emph{gustsfc}).
We consider a 24-hour period sampled at 1-hour increments over an equiangular spherical grid sampled at 1° increments.\footnote{Specifically the period from 00:00 to 23:59 of May 9, 2022. All times $t$ given in the text are in hours relative to the start of the period.}
We generate a spherical mesh from the 65160 locations of the grid by computing their convex hull with the Qhull software~\cite{barber1996quickhull}, for a total of 193320 edges.

\paragraph{Results} We train a conditional INR $f_\theta(\e_i, t)$ at the given spatial and temporal resolution, for $t = 0, \dots, 23$. Then, we evaluate its predictions on a high-resolution spherical mesh with 258480 nodes and 773280 edges (obtained with the same procedure of Section~\ref{sec:transferability_of_generalised_INRs}), and at time increments of 30 minutes.
We report in Figure~\ref{fig:weather_results} the prediction of the INR for a random test time, on both the original mesh and the high-resolution one.
We refer the reader to the animated version of the figure available in the supplementary material for a better appreciation of the results.
Overall, our results confirm that generalised INRs can model complex real-world signals on non-Euclidean domains, learning a realistic continuous approximation of the signals from low-resolution samples.

\section{Conclusion}
\label{sec:conclusion}

We have presented the problem of learning generalised implicit neural representations for signals on non-Euclidean domains. 
Our method learns to map a spectral embedding of the domain to the value of the signal, without relying on a choice of coordinate system. We have shown applications of our method on biological, social, and meteorological data, highlighting the potential usefulness of such INRs in a wide range of scientific fields.
We hope that future work will explore more real-world applications of our method.

\bibliographystyle{plainnat}
\bibliography{references}

\clearpage
\appendix
\section{Solving the Poisson equation with generalised INRs}

One interesting application of INRs is to train them using the derivatives of the target signal as supervision. This idea, which was introduced by~\citet{sitzmann2020implicit}, can also be applied to the generalised case. 

Specifically, in this experiment we consider the bunny reaction-diffusion texture and train a generalised INR to minimise:
\begin{equation}
    \cL = \| \L \f - \L \f_\theta \|,
\end{equation}

where $\L$ is the graph Laplacian, $\f = [ f(v_1), \dots, f(v_n) ]^\top \in \bR^n$ is the target graph signal, and $\f_\theta = [ f_\theta(\e_1), \dots, f_\theta(\e_n) ]^\top \in \bR^n$ is the signal predicted by the INR. 

\begin{figure}[!h]
    \centering
    \begin{subfigure}[T]{0.3\textwidth}
        \includegraphics[width=\textwidth]{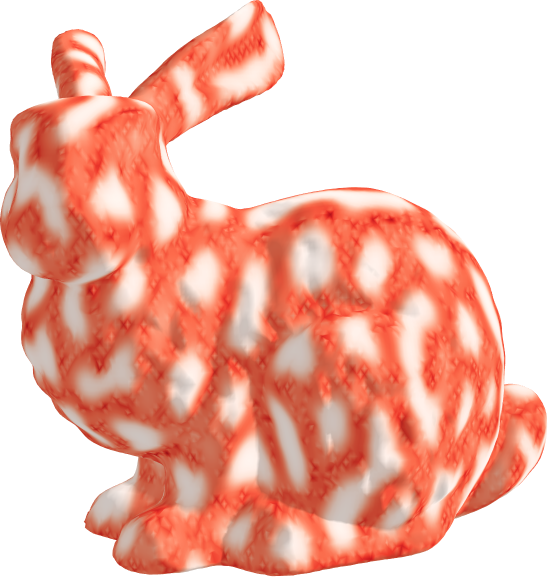}
        \caption{Original}
    \end{subfigure}
    \begin{subfigure}[T]{0.3\textwidth}
        \includegraphics[width=\textwidth]{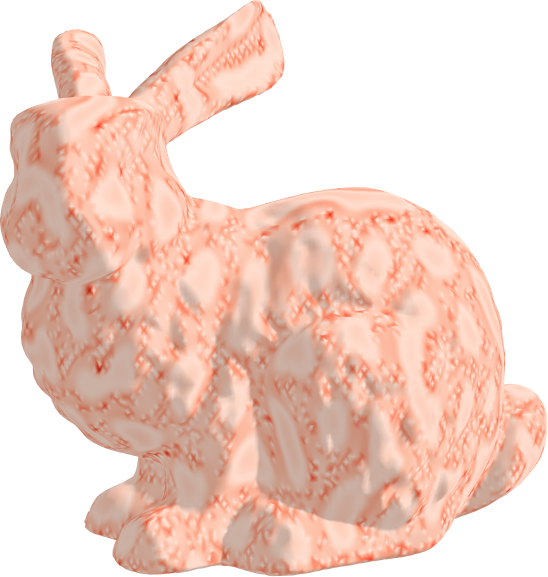}
        \caption{Laplacian}
    \end{subfigure}
    \begin{subfigure}[T]{0.3\textwidth}
        \includegraphics[width=\textwidth]{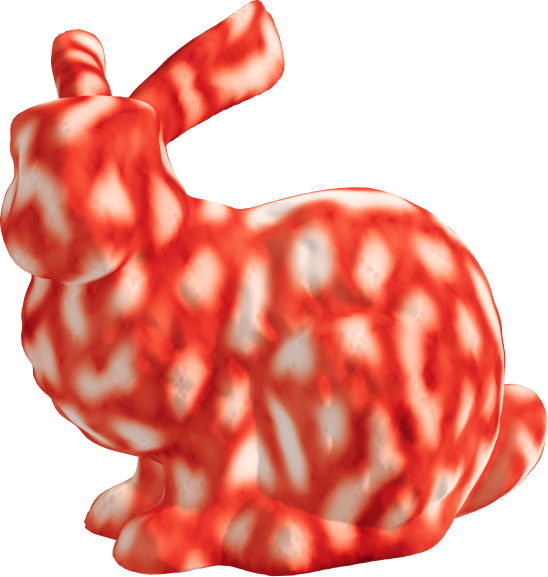}
        \caption{Reconstructed}
    \end{subfigure}
    \caption{The original signal, its laplacian, and its reconstructed version for the bunny reaction-diffusion texture.}
    \label{fig:poisson}
\end{figure}

We report in Figure~\ref{fig:poisson} the original texture, its Laplacian, and the reconstructed signal, as predicted by the INR. We see that the model is able to correctly reconstruct the signal, although with some oversmoothing. 

\section{Changes in the experimental setting}

\paragraph{Spectral embeddings} In the transferability experiments, we observed that changes in the high-frequency eigenvectors of different graph realizations of $\cT$ caused the INR to perform poorly at test time. 

For this reason, we used smaller spectral embeddings of size $k=3$ for the SBM experiment (which was enough to highlight the community structure of the graphs) and $k=7$ for the super-resolution experiment with the bunny (which we found by trial and error balancing the speed of convergence and the final performance).

In the weather experiment, the equiangular grid returned by GFS has a higher point density at the poles.
This non-uniformity has the effect that the first 66 eigenvectors are constant everywhere except at a very concentrated region around the poles, and they tend to change a lot when sampling more points (making it harder to transfer the trained INR to the high-resolution graph).
To improve stability, we removed the first 66 almost-trivial eigenvectors and trained the INR only on the remaining 34.

\paragraph{Architecture} We also observed that the SIREN multi-layer perceptron was sometimes too sensitive to small changes in the spectral embeddings. 
For this reason, we searched for a different architecture that would make the INR more transferable to different graphs. 

The alternative architecture has the following characteristics: 
\begin{itemize}
    \item ReLU activations;
    \item Uniformly distributed initial weights;
    \item 8 layers instead of 6 (we searched for the best value in $[4, 10]$);
    \item Learning rate of $10^{-3}$ (the default of $10^{-4}$ was not enough for the model to converge);
\end{itemize}

Note that we still kept the SIREN architecture for all other experiments as it exhibited better speed of convergence and lowest training loss compared to the alternative architecture. 

\section{Training details}

In the second experiment of Section 4.1, we train the models using the default setting. 
We create random training, validation and test splits with a proportion of 80\%, 10\%, and 10\% of the node set. 
We train the models to convergence, interrupting training if the validation loss does not improve for 1000 steps (we also lower the learning rate annealing to have a patience of 500 steps).
To compare the different activations, we simply swap the activation function and initialisation schemes, leaving everything else the same. 

\section{Alternative normalisation of the Laplacian}

Using different Laplacian normalisation schemes for computing the spectral embeddings did not yield significant differences in performance when training generalised INRs. 

\begin{figure}
    \centering
    \includegraphics[width=0.7\textwidth]{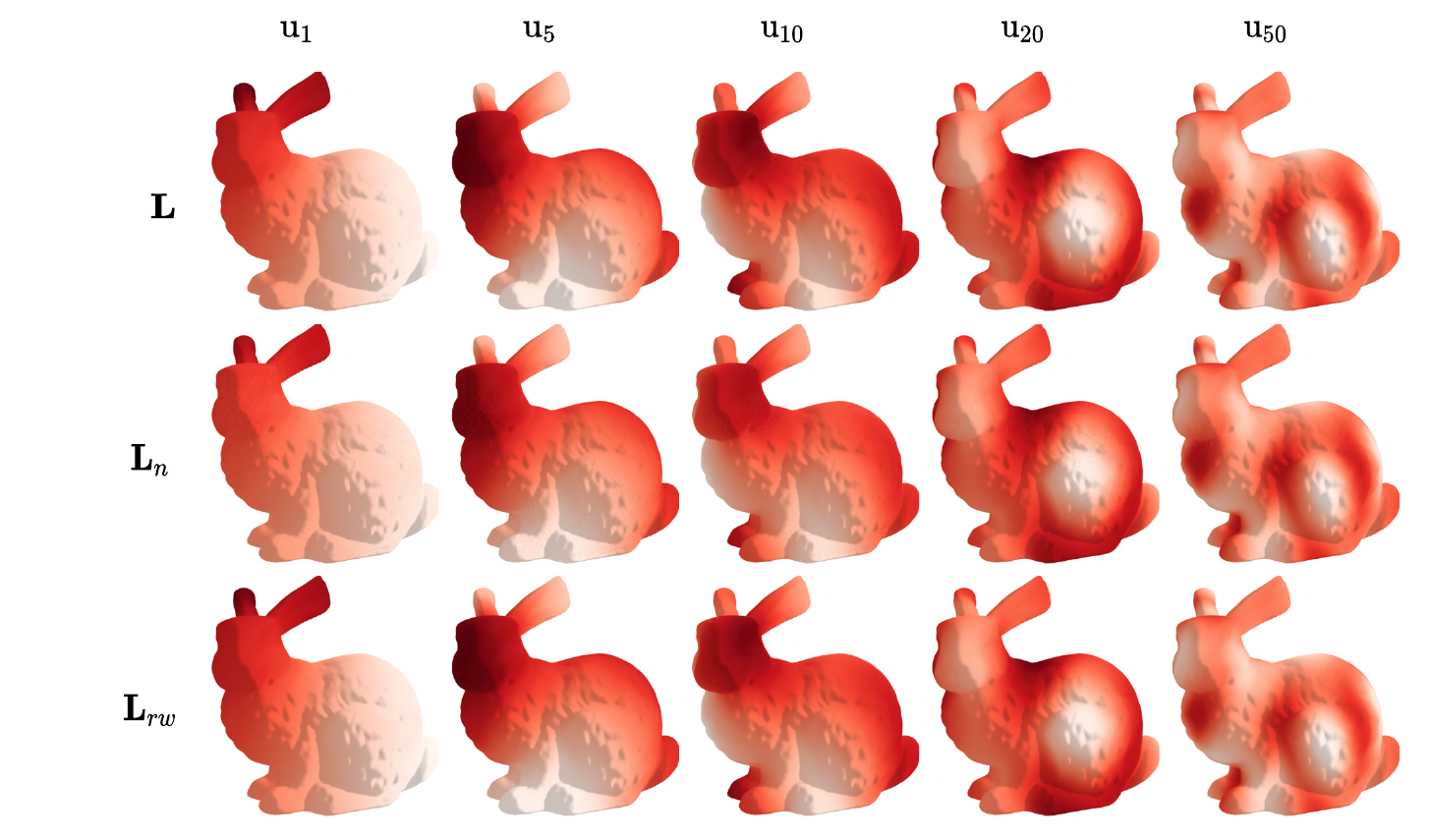}
    \caption{Eigenvectors $\u_k$ for $k = [1, 5, 10, 20, 50]$ for different Laplacian normalisation schemes, plotted on the Bunny mesh. Colours indicate intensity and colour scales are not shared between different rows.}
    \label{fig:laplacian_comparison}
\end{figure}

While the magnitude of individual embeddings will change across normalisations, they are qualitative similar and provide a reasonable encoding for a node's position. For visualizing this equivalence, we report in Figure~\ref{fig:laplacian_comparison} eigenvectors $\u_k$ for $k = [1, 5, 10, 20, 50]$ for three types of Laplacian: the combinatorial Laplacian, $\L = \D - \A$; the symmetrically normalised Laplacian, $\L_{n} = \D^{-1/2}\L\D^{-1/2}$; and the random walk normalised Laplacian $\L_{rw} = \D^{-1}\L$.

\clearpage
\section*{Checklist}
\begin{enumerate}

\item For all authors...
\begin{enumerate}
    \item Do the main claims made in the abstract and introduction accurately reflect the paper's contributions and scope?
    \answerYes{}
    \item Did you describe the limitations of your work?
    \answerYes{See Sections \ref{sec:method} and \ref{sec:experiments}}
    \item Did you discuss any potential negative societal impacts of your work?
    \answerNA{}
    \item Have you read the ethics review guidelines and ensured that your paper conforms to them?
    \answerYes{}
\end{enumerate}

\item If you are including theoretical results...
\begin{enumerate}
    \item Did you state the full set of assumptions of all theoretical results?
    \answerNA{}
    \item Did you include complete proofs of all theoretical results?
    \answerNA{}
\end{enumerate}

\item If you ran experiments...
\begin{enumerate}
    \item Did you include the code, data, and instructions needed to reproduce the main experimental results (either in the supplemental material or as a URL)?
    \answerYes{}
    \item Did you specify all the training details (e.g., data splits, hyperparameters, how they were chosen)?
    \answerYes{At the beginning of Section~\ref{sec:experiments} and in Section~\ref{sec:learning_generalised_INRs}}
    \item Did you report error bars (e.g., with respect to the random seed after running experiments multiple times)?
    \answerNA{}
    \item Did you include the total amount of compute and the type of resources used (e.g., type of GPUs, internal cluster, or cloud provider)?
    \answerYes{}
\end{enumerate}

\item If you are using existing assets (e.g., code, data, models) or curating/releasing new assets...
\begin{enumerate}
    \item If your work uses existing assets, did you cite the creators?
    \answerYes{}
    \item Did you mention the license of the assets?
    \answerNA{}
    \item Did you include any new assets either in the supplemental material or as a URL?
    \answerYes{}
    \item Did you discuss whether and how consent was obtained from people whose data you're using/curating?
    \answerNA{}
    \item Did you discuss whether the data you are using/curating contains personally identifiable information or offensive content?
    \answerNA{}
\end{enumerate}

\item If you used crowdsourcing or conducted research with human subjects...
\begin{enumerate}
    \item Did you include the full text of instructions given to participants and screenshots, if applicable?
    \answerNA{}
    \item Did you describe any potential participant risks, with links to Institutional Review Board (IRB) approvals, if applicable?
    \answerNA{}
    \item Did you include the estimated hourly wage paid to participants and the total amount spent on participant compensation?
    \answerNA{}
\end{enumerate}
\end{enumerate}

\end{document}